\documentclass[runningheads]{llncs}
\usepackage{graphicx}
\usepackage{float}
\begin{document}
\title{Toward Automatic Interpretation of 3D Plots}
%
%
\author{Laura E. Brandt\inst{1} \ and William T. Freeman\inst{1,2}\\\email{\{lebrandt,billf\}@mit.edu}}
%
\authorrunning{L.E. Brandt and W.T. Freeman}
%
\institute{MIT CSAIL, Cambridge MA, USA \and The NSF AI Institute for Artificial Intelligence and Fundamental Interactions}
%
\maketitle              
\begin{abstract}
This paper explores the challenge of teaching a machine how to reverse-engineer the grid-marked surfaces used to represent data in 3D surface plots of two-variable functions. These are common in scientific and economic publications; and humans can often interpret them with ease, quickly gleaning general shape and curvature information from the simple collection of curves. While machines have no such visual intuition, they do have the potential to accurately extract the more detailed quantitative data that guided the surface’s construction. We approach this problem by synthesizing a new dataset of 3D grid-marked surfaces (SurfaceGrid) and training a deep neural net to estimate their shape. Our algorithm successfully recovers shape information from synthetic 3D surface plots that have had axes and shading information removed, been rendered with a variety of grid types, and viewed from a range of viewpoints.

\end{abstract}
%
%
%
\section{Introduction}

\begin{figure}[t]
\begin{center}
   \includegraphics[width=\linewidth]{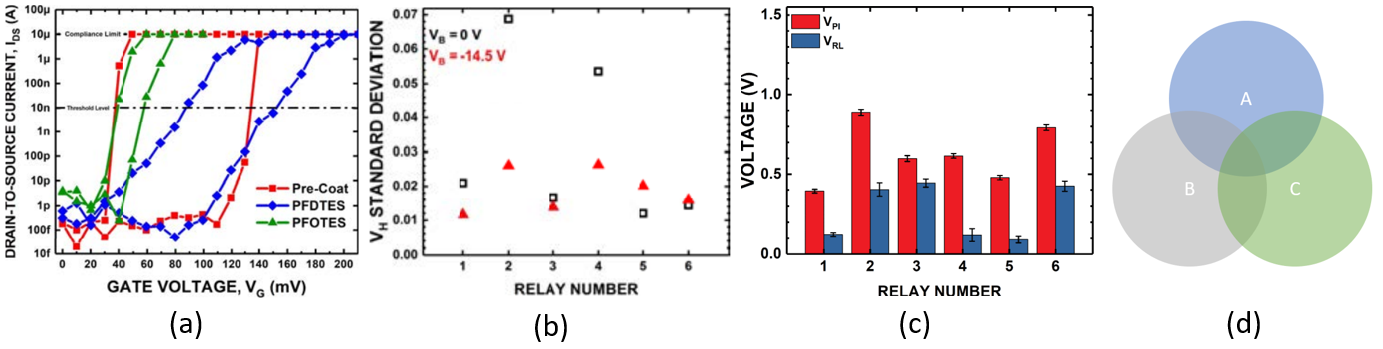}
\end{center}
   \caption{\textbf{2D charts.} Tools and algorithms exist to extract data from 2D charts like (a-c) experimental results from a micro-switch study \cite{Osoba}, and (d) Venn diagrams. No such methods exist for 3D charts like the surface plot in Figure \ref{fig:3dSurfaces}.}
\label{fig:2dPlots}
\end{figure}

\begin{figure}[t]
\begin{center}
   \includegraphics[width=\linewidth]{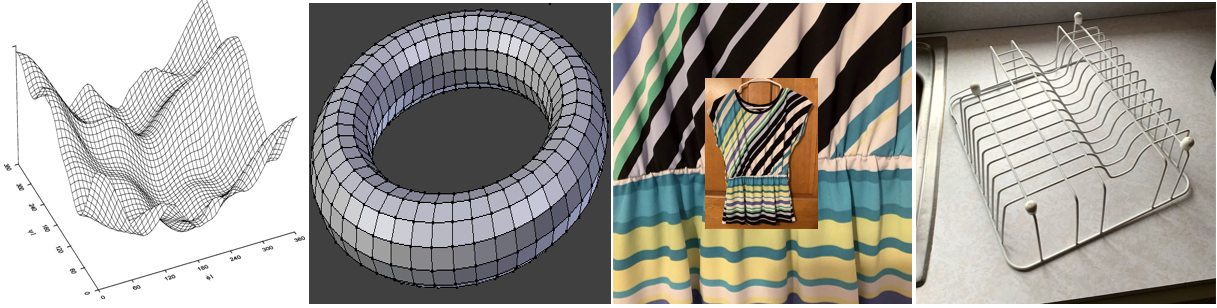}
\end{center}
   \caption{\textbf{Grid-marked surfaces} are common both in technical publications and everyday life, and humans can easily interpret shapes ranging from simple toroids to ripples in fabric. This paper works toward the automated interpretation (by computer) of 3D plots like the potential energy surface at left \cite{Zamarbide}, using neural models that might someday extend to the analysis of more general grid-marked surfaces such as this torus, shirt, and dish rack.}
\label{fig:3dSurfaces}
\end{figure}

Suppose you encounter a chart in a published paper and want to automate the interpretation of the data it represents so you can use it in your own research. If the chart were a 2D one like those shown in Figure \ref{fig:2dPlots} (\textit{e.g.} a pie chart or a scatter plot), you would be able to use any of an array of available tools like \cite{PlotDigitizer,Engauge} to recover the data yourself. However if the chart were instead a 3D surface plot like that shown in Figure \ref{fig:3dSurfaces}, you would be out of luck. Humans have the ability to easily perceive qualitative information such as shape and curvature from grid-marked surfaces like these, but to recover quantitative data from a surface plot by hand would be an arduous (and likely doomed) task. Machines on the other hand lack our visual intuition about surface contours. But properly designed they have the potential to do something humans cannot: reverse-engineer the surface to recover the original data that created it. This work explores the challenge of teaching a machine how to reverse-engineer 3D grid-marked surfaces. This visual perception task is of inherent interest, and supports the goal of recovering data from published 3D plots.

Internet search engine providers like Google and Microsoft have interest in tools that can look up and recover data from published sources. Motivated by the amount of information stored in tables on the Internet and in documents, Google AI recently published the TAPAS approach to automated table parsing, which allows the user to query published tables with questions phrased in natural language \cite{TAPAS1,TAPAS2}. An aim of such research is to improve user ability to comb the Internet or other large databases for published data based on their understanding of its content rather than a lucky guess of its associated keywords, and there is demand among scientists for tools to help them find published figures and papers based on their data content \cite{SourceData,CiteSeerX}. A model able to reverse-engineer published 3D plots would enable the vast digital library of technical publications containing such figures to be automatically processed and indexed based on the data they present, improving the quality and efficiency of scientific literature surveys.

There exist several kinds of 3D plots in published literature. This paper is concerned specifically with plots of two-variable functions like the potential energy surface shown in Figure \ref{fig:3dSurfaces}. To interpret such a figure, an interpreter must be able to do several things. It must be able to detect the figure in the first place. It has to segregate that figure into labels, axes, and the visual representation of the data (that is, the grid-marked deformed surface of the plot). Then it must reverse-engineer that visual representation to recover the data that was used to produce it. This involves perceiving the shape of these grid-marked surfaces, determining the perspective from which the surface plot was projected when the figure was rendered, undoing that perspective projection to recover the original data as a function $z(x,y)$, and calibrating that data using the axes and annotations earlier extracted from the figure.

This work is focused on the specific task of recovering shape information from grid-marked surfaces akin to 3D surface plots with axes removed. Figure detection, segregation, de-projection, and calibration are discussed further in the concluding section of this paper. Marr, Stevens, and others \cite{Marr,Stevens} have approached the more general problem of shape-from-surface contours by making a variety of assumptions about surface symmetries and constraints (\textit{e.g.} that surfaces are locally cylindrical \cite{Marr,Stevens}). These are conditions that often do not hold in 3D surface plots, and we instead approach our shape recovery problem with an artificial neural net.


The main contributions of this work are
\begin{enumerate}
    \item A deep neural net trained to recover shape information from 3D surface plots without axes or shading information, synthesized with a variety of grid types and viewed from a range of viewpoints.
    \item SurfaceGrid: a 98.6k-image dataset of such 3D plot surfaces, synthesized side-by-side with their corresponding depth maps and underlying data.
    \item An experimental analysis of the neural model and training method, including an ablation study and performance evaluation.
\end{enumerate}


\section{Related work}

\textbf{Automatic chart interpretation} is an active area of research and there exist a large number of semi-automatic tools like \cite{PlotDigitizer,Engauge,GraphGrabber,WebPlotDigitizer}, often called ``plot digitizers", that can extract data from published 2D charts like those shown in Figure \ref{fig:2dPlots} (bars, pies, \textit{etc.}). However, these tools generally require that the human user manually perform key steps like axis calibration or feature identification. This requirement of human input for every image limits the application of such tools to the occasional reverse-engineering of single figures or to applications for which the number of man-hours required is deemed worth it.

Work has been done toward fully automating the processes of 2D chart interpretation. Yanping Zhou and Tan \cite{Zhou} used edge tracing and a Hough transform-based technique to extract bars from bar charts. Huang \textit{et al.} and Liu \textit{et al.} \cite{Huang1,Huang2} used vectorized edge maps to extract features from bar, 2D line, and other 2D charts. Huang and Tan \cite{Huang3}, Lu \textit{et al.} \cite{Lu1}, and Shao and Futrelle \cite{Shao} used these kinds of features to classify chart types, and Lu \textit{et al.} \cite{Lu2} used them to extract and re-plot data from smooth 2D line plots. Prasad \textit{et al.} \cite{Prasad} measured similarity between the scale-invariant feature transform (SIFT) \cite{Lowe} and histograms of oriented gradients (HOG) \cite{Dalal} and used it to train a support vector machine (SVM) \cite{SVM} to classify chart types. ReVision \cite{Savva} used a combination of feature identification and patch clustering to not only classify figures but also reverse-engineer their data to enable re-visualization to other chart formats, and Jung \textit{et al.} and Dai \textit{et al.} \cite{Dai,Jung} similarly trained classifiers to categorize published charts in addition to extracting features and text. Last year, Fangfang Zhou \textit{et al.} \cite{Fangfang} took a fully neural network-based approach to interpreting bar charts, using a Faster-RCNN \cite{FRCNN} to locate and classify textual chart elements, and an attentional encoder-decoder to extract numerical information. To our knowledge the prior work focuses entirely on 2D charts, leaving the problem of interpreting 3D surface plots like that in Figure \ref{fig:3dSurfaces} unaddressed.

\textbf{Shape-from-X.} Many kinds of cues have been used to teach machines to reconstruct shape from images including symmetry \cite{Francois,Mukherjee}, shading \cite{Horn,Zhang}, stereo \cite{Kim,DeVries} and multi-view \cite{Godard}, motion \cite{Faugeras,Ummenhofer,Tinghui}, silhouette \cite{Koenderink}, and texture \cite{Witkin}. Our work was particularly inspired by Marr's discussion of the human ability to perceive shape from from simple grid- and line- marked surfaces \cite{Marr}, which cited Stevens as an example of a computational attempt to reconstruct shape from surface contour cues by making the simplifying assumption that all instances of curvature are approximately cylindrical \cite{Stevens}. Weiss \cite{Weiss} proposed an alternative algorithm that modeled surface contours as springs with energy to be minimized in solving for surface shape, while Knill \cite{Knill} proposed that image contours ought be interpreted with a human-inspired geodesic constraint, and Ulupinar and Nevatia \cite{Ulupinar} introduced a method for reconstructing the shape of generalized cylinders from median lines by assuming generic viewpoint \cite{Freeman} and certain symmetries. Mamassian and Landy \cite{Mamassian} conducted a psychophysical study of the stability of human interpretations of simple grid-marked surfaces, and modeled their process as a simple Bayesian observer. More recently, Pumarola \textit{et al.} \cite{Pumarola} used a convolutional neural net to estimate the meshgrids of synthesized deformed surfaces ``made" with various materials (fabric, metal, \textit{etc.}), and Li \textit{et al.} \cite{BendSketch} introduced a tool (BendSketch) for sketch-based 3D model design capable of generating 3D models from sketches using an input collection of digital pen annotations from pre-defined categories of surface contours.


\section{Method}


\subsection{Network architecture}

\begin{figure}[t]
\begin{center}
    \includegraphics[width=\linewidth]{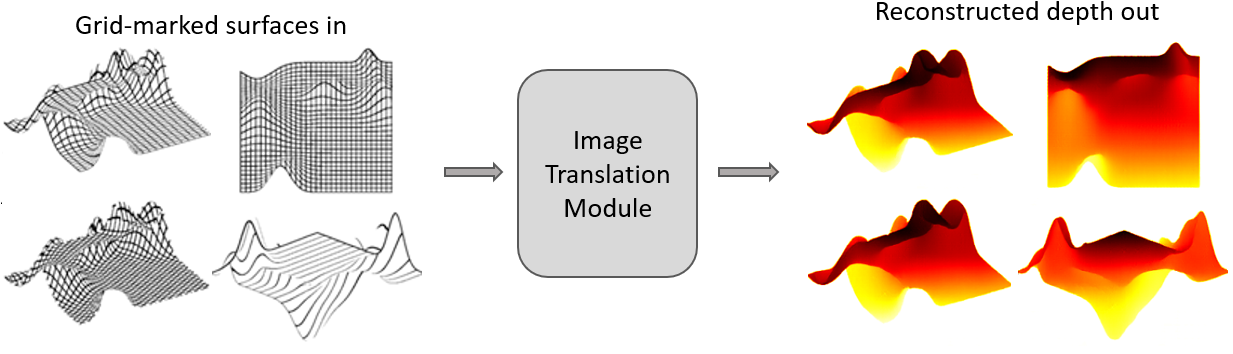}
\end{center}
   \caption{\textbf{Our Final model} used an image translation module \cite{P2P} to reverse-engineer grid-marked surfaces and recover depth maps representing their underlying data.}
\label{fig:final}
\end{figure}

The task of reverse-engineering shape information from a picture of a 3D surface plot can be approached as an image translation task (Figure \ref{fig:final}), and we did so using the Pix2Pix neural image translation module \cite{P2P}. Pix2Pix is a conditional generative adversarial network (cGAN) consisting of a UNet \cite{ronneberger2015u} generator that learns to map (in our case) from an input surface plot to an output ``fake" depth map, and a PatchGAN \cite{P2P} discriminator that takes a depth map as input and learns to identify whether it is real or merely a generated estimate by looking at small patches of local structure. The discriminator provides feedback to the generator during training, so that the later learns simultaneously to ``fool" the discriminator and to minimize the mean absolute error between estimated and true depth map. We refer readers to Isola \textit{et al.}'s paper for further detail \cite{P2P}.

\subsection{The SurfaceGrid dataset}

A sizable dataset is required to train a neural net, and the creation of a good one is a challenge. While images of grid-marked surfaces could in principle be gathered from ``the wild", only rarely is truth data available. This necessitated the side-by-side synthesis of a new dataset: SurfaceGrid.

\begin{figure}[t]
\begin{center}
   \includegraphics[width=0.75\linewidth]{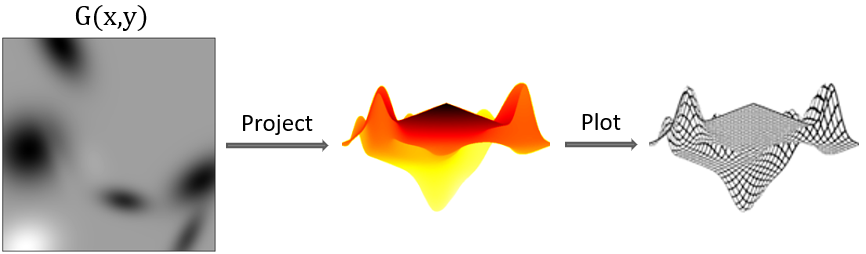}
\end{center}
   \caption{\textbf{Dataset generation} was a three-step process. First multivariable functions $G(x,y)$ were created by summing synthesized Gaussians. These were then projected as depth maps and plotted from specified camera viewpoints. These projected plot-depth map pairs were used to train our model.}
\label{fig:generation}
\end{figure}

We produced a dataset of what are effectively mathematical surface plots without axes or shading, paired with depth maps. This allowed us to use the following 3-step procedure for synthesizing a large dataset (Figure \ref{fig:generation}):
\begin{enumerate}
    \item randomly generate a set of multivariable functions $G(x,y)$,
    \item project these functions as depth maps from camera viewpoints of choice, and
    \item plot these functions with a variety of grid parameters, projected from the same viewpoints as in Step 2.
\end{enumerate}

\begin{figure}[t]
\begin{center}
   \includegraphics[width=0.75\linewidth]{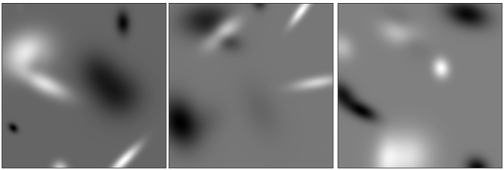}
\end{center}
   \caption{\textbf{Multivariable functions} $G(x,y)$ were generated for our new dataset by summing Gaussians with randomly-synthesized parameters. Gaussians are radial basis functions, and sums of them can be used to model any two-variable function. These $G(x,y)$ were subsequently projected as depth maps and plotted.}
\label{fig:datasetFunctions}
\end{figure}

In the first step we randomly generated $1 \leq N \leq 10$ Gaussians $g_i(x,y)$ with standard deviations $8 \leq \sigma_x, \sigma_y \leq 512$ pixels, rotated by a random angle in the $xy$ plane, and shifted to a random mean $\mu$ anywhere on the domain $0 \leq x, y \leq 512$ pixels. The individual Gaussian amplitudes were chosen to normalize their areas, and could be positive or negative with equal probability. These $N$ Gaussians $g_i(x,y)$ were then summed to produce the final multivariable function $G(x,y) = \sum_{i = 1}^N g_i(x,y)$. Figure \ref{fig:datasetFunctions} shows examples.

In the second step we took these functions and computed their corresponding depth maps as projected from a set of camera viewpoints chosen to cover an octant of space. Specifically, in spherical coordinates $(r, \theta, \phi)$ we chose a camera distance $r = 1024$ pixels from the origin and azimuth and elevation angles $\theta, \phi \in [0, 30, 60]\deg$. We generated an additional viewpoint of $(\theta, \phi) = (45, 22.5)\deg$.

In the third and final step we generated grid-marked surface plots of these functions using lines $\delta x = \delta y = 3$ pixels wide and spaced $\Delta x = \Delta y \in [17, 25, 34,$ $42]$ pixels apart for effective sampling resolutions of $\Delta X \equiv \Delta x + \delta x = \Delta Y \equiv \Delta y + \delta y \in [20, 28, 37, 45]$ pixels. These numbers were chosen somewhat arbitrarily based on a human judgement of what looked ``reasonable but different", and it is worth noting here that the $\Delta x = \Delta y = 42$ pixel case was often guilty of undersampling the data. We also produced line- (not grid-) marked surfaces, one type of grid with different line spacings in the two directions ($\Delta X = 20 \neq \Delta Y = 28$ pixels), and grids rotated by azimuthal angles $\theta_g \in [30, 50, 60] \deg$.

\begin{figure*}
\begin{center}
\includegraphics[width=\linewidth]{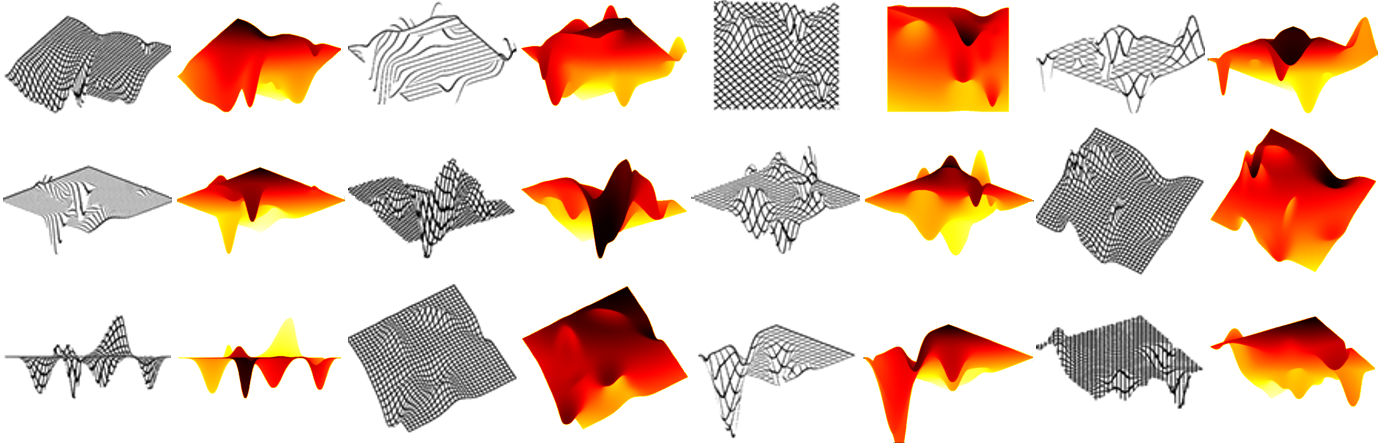}
\end{center}
   \caption{\textbf{The SurfaceGrid dataset} is our new 98.6k-image dataset. It contains synthesized binary 3D surfaces marked with grids and lines of varying resolutions, viewed from a multitude of camera positions, and paired with their associated ground truth depth maps. The underlying multivariable functions are also included. The dataset is available online at https://people.csail.mit.edu/lebrandt.}
\label{fig:dataset}
\end{figure*}

Overall, 76.6k images of grid-marked surfaces, 20k corresponding projected depth maps, and 2000 underlying multivariable functions were generated for a total of 98.6k images in this dataset (Figure \ref{fig:dataset}). The grid-marked surfaces shown here and throughout the paper are inverted for printing purposes from the raw data in the dataset which is in inverse (white-on-black). All images are single-channel, 512x512, and available online at https://people.csail.mit.edu/lebrandt.

\subsection{Training details}

Published 3D surface plots come in many forms; the data being plotted, surface marking resolutions, and rendering viewpoints all vary greatly. In this work we used sums of radial basis functions (Gaussians) in an effort to efficiently cover the enormous domain of published data with a small number of parameters. Any two-variable function can be modeled by summed Gaussians, if a sufficient number are used. We needed to do similarly with the domains of surface markings and viewpoints.

We could have trained our model on examples rendered with randomly-generated markings and viewpoints, but were concerned that it would over-train in a way difficult to detect via experiment. When training examples are randomly-generated, randomly-generated test examples are very likely to have a ``near neighbour" in the training set that is near enough to result in a low mean error, but far enough that the rules being used to reverse-engineer the surface's shape are not general. We therefore built up our training set by adding \textit{specific} categories of grid-marked surfaces to a very narrow \textit{Baseline} training set. By training models and testing as we went (using test parameters very different from training examples, \textit{e.g.} viewpoint shifted from 30 to 60 deg azimuth), we were able to see how general the rules learned by the model were. We will discuss this further in Section \ref{ablation}.



We used the Pix2Pix architecture introduced in \cite{P2P} and implemented for PyTorch in \cite{CycleGAN}, using an Adam \cite{Adam} optimizer with learning rate = 0.0002 and momentum parameters $\beta_1 = 0.5$ and $\beta_2 = 0.999$. The training set size was 1.8k image pairs for all models, created via a shuffled sampling of surfaces corresponding to functions 0-1799. Mean absolute error between reconstruction and ground truth was computed over a 100-image pair validation set (functions 1800-1899). Training was stopped (and the model rolled back) when the validation error had been increasing continuously for 50 epochs. Our \textit{Final} model trained for 274 epochs.


\section{Experiments}
For all experiments we used mean-squared relative error (MSRE) between reconstruction and ground truth as our metric, averaged over 100-image pair test sets (functions 1900-1999).

\subsection{Final model performance}\label{finalPerformance}

We applied our \textit{Final} model both to a general (\textit{Full}) test set of 3D surfaces marked with a variety of grid types, and to six more narrowly-scoped test sets that isolated specific surface types. The narrowest \textit{Base} test set consisted solely of surfaces marked with 20x20 pixel grids and projected from camera viewpoints that had appeared in the training set. We excluded viewpoints with extreme (very high/low) elevation angles.

Five test sets added a single class of surfaces to the \textit{Base} set. The \textit{Resolutions} and \textit{Angled Grids} test sets added new grid resolutions and angles, respectively. The \textit{Viewpoints} dataset added surfaces viewed from new generic viewpoints. The \textit{Lines} dataset added surfaces marked with lines instead of grids. Finally, the \textit{Accidental Viewpoints} dataset restored the previously-excluded viewpoints with extreme elevation angles. The \textit{Full} test set included images of all these types, including images that combined multiple of these effects. (\textit{E.g.} line-marked surfaces viewed from new viewpoints, or surfaces marked with angled lines.)

\begin{figure*}
\begin{center}
\includegraphics[width=\linewidth]{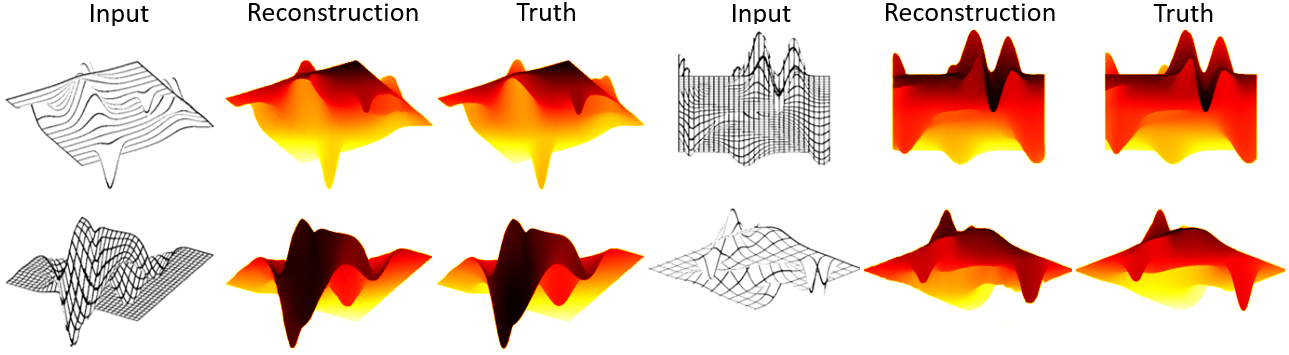}
\end{center}
   \caption{\textbf{Final model performance} when applied to various input grid- and line- marked images. It performed well on a variety of grid resolutions, angles, and generic camera viewpoints. Top row inputs had similar parameters as during training. Bottom row inputs tested (L) new camera viewpoint and (R) new viewpoint and grid resolution.}
\label{fig:results}
\end{figure*}

\begin{figure}[t]
\begin{center}
   \includegraphics[width=\linewidth]{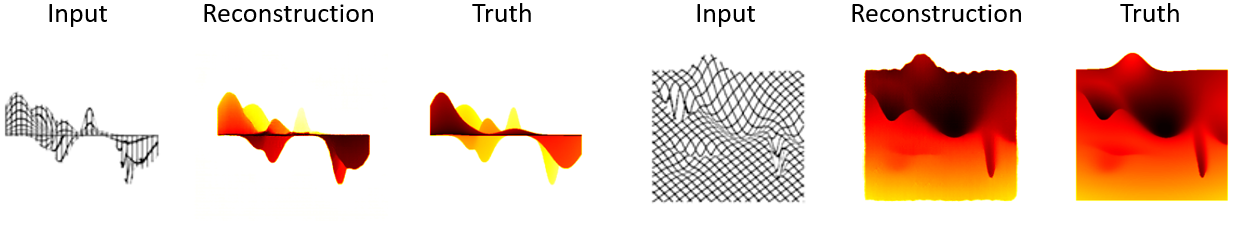}
\end{center}
   \caption{\textbf{Limitations.} The \textit{Final} model struggled to generalize to extreme-elevation viewpoints, especially when the grid was angled. There is an inherent depth ambiguity in edge-on projections at low elevations because occlusion cues are lacking. At high elevations, curvature cues are lost and, in the case of angled grids, plot boundaries interpreted as jagged. Note that data is rarely plotted from such extreme elevations.}
\label{fig:limitations}
\end{figure}

Table \ref{tab:results} reports the MSRE for our \textit{Final} model when applied to these test sets, and Figure \ref{fig:results} shows characteristic results. The \textit{Final} model generalized well to viewpoints and line/grid resolutions not seen during training, though it struggled with surfaces viewed from extreme elevation angles (Figure \ref{fig:limitations}).

\begin{table}
\caption{\textbf{Final model performance} on narrowly-scoped test sets. The \textit{Final} model generalized well to new grid resolutions and angles, line-marked surfaces, and generic viewpoints. It struggled with surface plots viewed from extreme elevation angles (\textit{Acc. Viewpoints}), but overall generalized well to the \textit{Full} test set (last row). None of the reconstruction errors exceeded 0.5\% MSRE.}
\begin{center}
\begin{tabular}{c c}
\hline
\textbf{Test Set} & \textbf{MSRE (x$10^{-2}$)} \\ \hline
Base              & 0.106               \\ \hline
Resolutions               & 0.107               \\
Angled Grids               & 0.124               \\
Lines               & 0.213               \\
Viewpoints               & 0.281               \\
Acc. Viewpoints              & 0.424               \\ \hline
Full              & 0.327               \\ \hline
\end{tabular}
\end{center}
\label{tab:results}
\end{table}

\subsection{Impact of training set components on performance}\label{ablation}

\begin{figure}[t]
\begin{center}
   \includegraphics[width=0.65\linewidth]{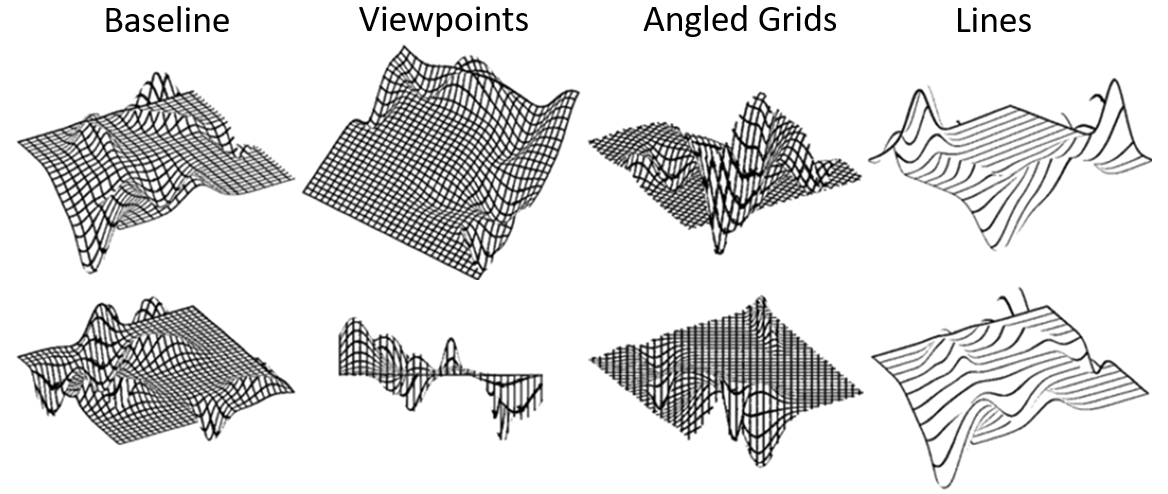}
\end{center}
   \caption{\textbf{Reduced training sets.} Example images from the four reduced training sets discussed in Section \ref{ablation}. \textit{Baseline} contained only 20x20 pixel grids viewed from a single viewpoint. \textit{Viewpoints} added additional general viewpoints. \textit{Angled Grids} and \textit{Lines} added angled grids and line-marked surfaces, respectively, to the \textit{Viewpoints} dataset. The \textit{Final} model used a training set containing all of these image types.}
\label{fig:reducedTraining}
\end{figure}

We compared the performance of our \textit{Final} model with that of four variants trained on reduced datasets of increasing complexity (Figure \ref{fig:reducedTraining}). The na\"{i}ve \textit{Baseline} model was trained using only 20x20 grid-marked surfaces viewed from a single generic viewpoint of $\theta = \phi = 30\deg$. The \textit{Viewpoints} training set added general viewpoints $\theta, \phi \in [0,30,60]$ to the \textit{Baseline} training set (chosen to cover an octant of space). Elevation $\phi = 90\deg$ was neglected because grid-marked surfaces lack curvature cues when viewed directly from above, and azimuths $\theta \geq 90\deg$ were excluded because they have equivalent viewpoints in $0 \leq \theta < 90 \deg$. Adding these eight viewpoints to the dataset allowed our model to recover shape from arbitrary generic viewpoints, but it could not handle general grid resolutions.

We found that training on a single additional class of input images was sufficient: surfaces marked with \textit{Lines} (not grids) of a new spacing (we somewhat arbitrarily chose 37 pixels). Even though this class of training data was \textit{line}-marked, it enabled the model to generalize to completely new \textit{grid} resolutions. As an experiment we added surfaces with \textit{Angled Grids} that had been rotated by an azimuthal angle $\theta_g \in [30, 60]$, which enabled generalization to new grid angles but lowered overall performance. Angled grids are also quite rare in 3D surface plots. Figure \ref{fig:reducedTraining} shows examples from these reduced training sets.

We applied these models both to a general (\textit{Full}) test set of 3D surfaces marked with a variety of grid types, and to three more narrowly-scoped test sets. The \textit{Base} and \textit{Full} test sets were the same as in Section \ref{finalPerformance}. The \textit{General Grids} test set added new grid resolutions and angles, along with line-marked surfaces, to the \textit{Base} test set. Alternatively, the \textit{General Views} test set added new viewpoints without diversifying the surface markings themselves.

\begin{figure*}
\begin{center}
\includegraphics[width=\linewidth]{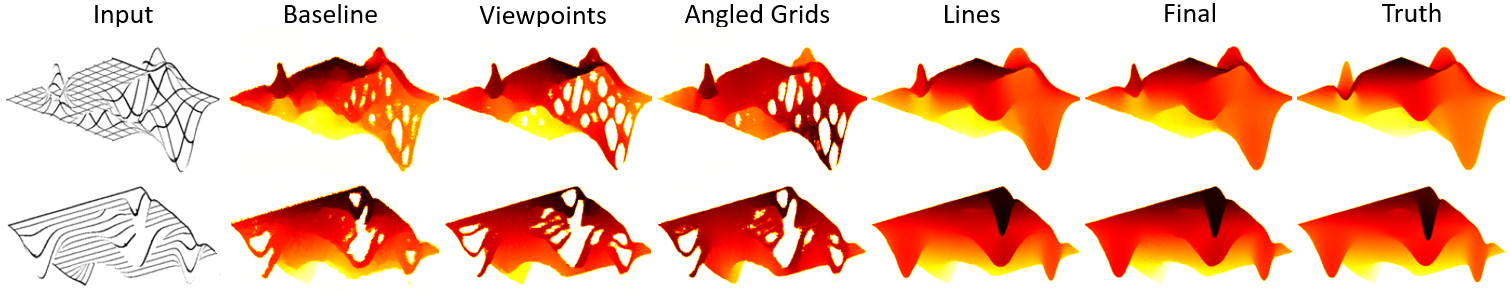}
\end{center}
   \caption{\textbf{Relative impact} of of different additions to the training set on reconstructions. The addition of line-marked surfaces to the training set was particularly impactful (third-to-last column). The bottom row input had similar parameters as during training. The top row input tested a new viewpoint and grid resolution.}
\label{fig:comparison}
\end{figure*}

Table \ref{tab:comparison} compares the MSRE of the reconstructions produced by these five models when applied to these four test sets. Table \ref{tab:impact} reports the relative impact of each addition to the training set on model performance, expressed as a percent improvement in MSRE on these four test sets. Figure \ref{fig:comparison} gives a few examples of the impact of these training set additions on the model's ability to generalize to new grid resolutions and viewpoints.

\begin{table}
\caption{\textbf{Impact} of different additions to the training set (down) on model performance when applied to specific types of images (across), reported as MSRE x$10^{-2}$. Our \textit{Final} model generalizes best to the \textit{Full} and \textit{General Grids} test sets, and performs within 2\% and 4\% of the best on the \textit{General Views} and \textit{Base} sets, respectively.}
\begin{center}
\begin{tabular}{l c c c c}
\hline
$\downarrow$ Training $\backslash$ Test $\rightarrow$ & Base & Gen. Grids & Gen. Views & Full \\ \hline
Baseline   & 0.237         & 0.804         & 0.723          & 0.951         \\
Viewpoints     & 0.127         & 1.43          & 0.385          & 1.18          \\
Angled Grids     & 0.109         & 1.49          & \textbf{0.346}          & 1.20          \\
Lines     & \textbf{0.102}         & 0.183         & 0.355          & 0.329         \\ \hline
Final      & 0.106         & \textbf{0.148}         & 0.353          & \textbf{0.327}         \\ \hline
\end{tabular}
\end{center}
\label{tab:comparison}
\end{table}

\begin{table}[H]
\caption{\textbf{Relative impact} of different additions to the training set (down) on model performance when applied to specific types of images (across), reported as a \% improvement in MSRE. Positive numbers are good and the largest improvements in each category are in bold. \textit{Angled Grids} were not particularly beneficial inclusions, while \textit{Lines} were particularly important for enabling generalization to new grid resolutions and improving overall performance on the \textit{Full} test set.}
\begin{center}
\begin{tabular}{l c c c c c}
\hline
$\downarrow$ Training $\backslash$ Test $\rightarrow$ & Base     & Gen. Grids     & Gen. Views    & Full     \\ \hline
Baseline & - & - & - & - \\
Viewpoints           & \textbf{46.4}  & -77.7 & \textbf{46.8} & -24.3 \\
Angled Grids           & 14.2 & -4.15 & 10.0   & -1.95 \\
Lines           & 19.7 & \textbf{87.2} & 7.80 & \textbf{72.2} \\ \hline
Final            & -3.92 & 19.3 & 0.564 & 0.608 \\ \hline
\end{tabular}
\end{center}
\label{tab:impact}
\end{table}

\subsection{Performance on real plots and wireframe models}

\begin{figure}[t]
\begin{center}
   \includegraphics[width=\linewidth]{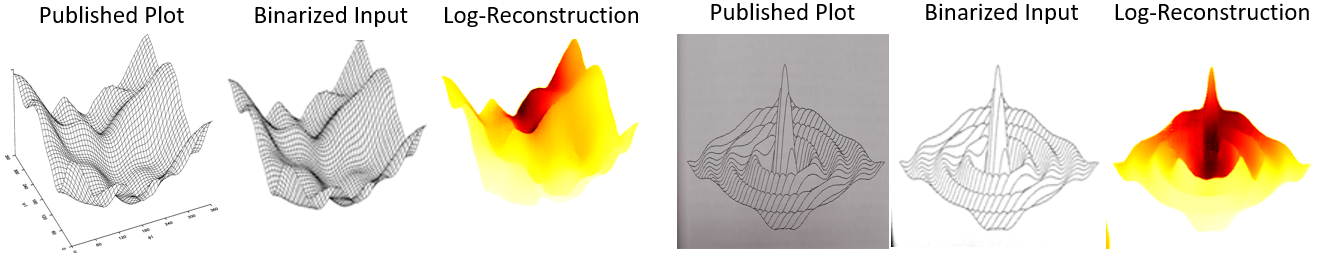}
\end{center}
   \caption{\textbf{Performance on real plots.} Our \textit{Final} model successfully extracted general shape and relative depth information from published plots taken from \cite{Marr,Zamarbide}, even though these plots would clearly require more than the 10 Gaussians summed in our synthetic dataset to model. Note that pre-processing was required to binarize the images and remove plot axes, and that the results are un-calibrated.}
\label{fig:realPlots}
\end{figure}

We applied our \textit{Final} model to real plots taken from publications \cite{Marr,Zamarbide}. In these cases we lacked truth data and so could not compute the MSRE for our reconstructions, but Figure \ref{fig:realPlots} shows that our model successfully extracted general shape and relative depth information from these plots.

\begin{figure}[t]
\begin{center}
   \includegraphics[width=\linewidth]{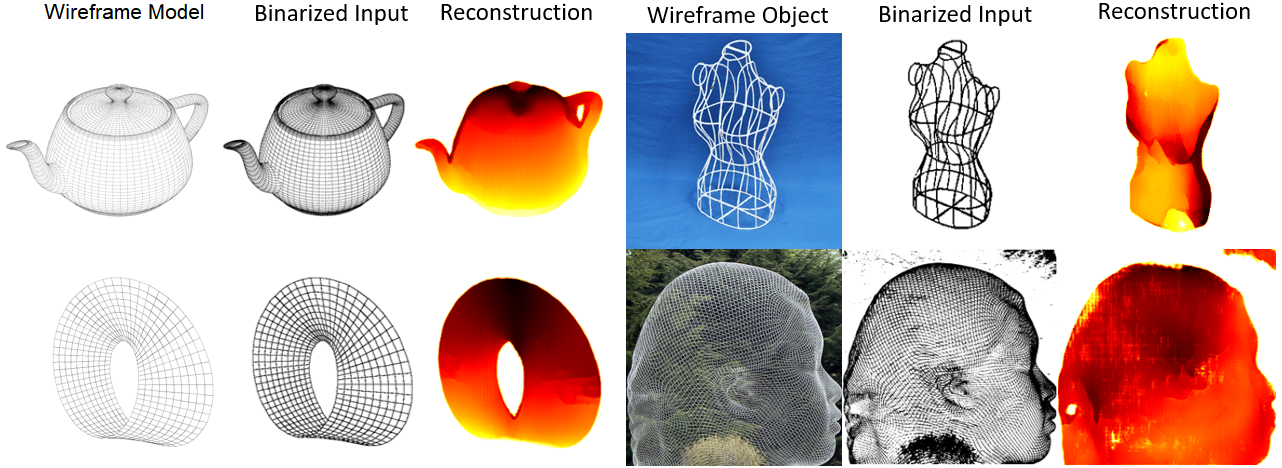}
\end{center}
   \caption{\textbf{Performance on wireframe models and objects.} We also applied our neural model to a variety of wireframe models from \cite{rm-hull} and real-world objects from \cite{sculpture,Vid2Curve} and found that despite clear differences between them and the training data, our model extracted meaningful shape information.}
\label{fig:wfModels}
\end{figure}

We also applied our neural model to wireframe models from \cite{rm-hull} and binarized images of physical wireframe objects from \cite{sculpture,Vid2Curve}. Figure \ref{fig:wfModels} shows that even for the object models, which look qualitatively very different from the synthetic training data, our neural model still manages to extract general relative depth information. Notice also that unlike the training data, the real-world wireframe objects do not have opaque surfaces between the surface contours, meaning that contours on front and back surface are equally visible. Despite this, our \textit{Final} model generates reasonable interpretations of entire sections of these images.

All of these surfaces would clearly require far more than ten Gaussian basis functions to fully model. They also featured different viewpoints and line spacings than the synthetic training data. The ability of our model to recover meaningful information from these images suggests that our carefully constructed training set successfully taught the neural net general rules for computing local shape from surface contours, suggesting high promise for use in automatic 3D plot and graphical model interpretation.


\section{Conclusion}

In order to reverse-engineer real published 3D plots, several problems remain to be solved. A front-end system to detect figures within documents and segregate them into axes, labels, and the plotted surface will be necessary. Additionally, the recovered depth maps must be calibrated and de-projected (to estimate the function $z(x,y)$) according to the axes. Since axes are orthogonal, the general procedure for de-projecting images can be simplified \cite{Tsai,Zhengyou}.

Our model successfully recovered depth maps from synthetic 3D surface plots without axis or shading information. It generalized to such surfaces rendered with line/grid resolutions and viewpoints that it had not encountered before (Figure \ref{fig:results}) with less than 0.5\% mean-squared relative error (Table \ref{tab:results}). When applied to real published plots, the quantitative data extracted by our model was un-calibrated; but as Figure \ref{fig:realPlots} clearly shows, it is possible to generate qualitative interpretations of the results. Our model was also able to extract meaningful information from pictures of wireframe models and objects (Figure \ref{fig:wfModels}).

We measured how different types of training set data (categorized by surface marking type/resolution and camera viewpoint) affected model performance and ability to generalize, and identified which ones were most (and least) important. Some kinds of data boosted performance far more than we expected; for example we found that training on \textit{line}-marked surfaces (in addition to a single-resolution grid-marked dataset) was responsible for a $>$70\% improvement in overall model performance (Table \ref{tab:impact}) \textit{and} the ability to generalize to new \textit{grid} resolutions (Figure \ref{fig:comparison}). We hope our study will help inform future training set design.

Automatic recovery of numerical data from published 3D plots is an exciting and difficult problem, and our work successfully demonstrates that a very simple neural net architecture is capable of accurately reverse-engineering 3D surface plots if it is carefully trained, the axes are identified and removed, and a calibration obtained. We establish a baseline of performance on this task, present SurfaceGrid (a new 98.6k-image dataset of grid-marked surfaces and their associated depth map truths, Figure \ref{fig:dataset}), and invite other researchers to contribute to this exciting effort.\newline

\noindent\textbf{Acknowledgement.} This work is supported by the National Science Foundation (NSF) under Cooperative Agreement PHY-2019786 (The NSF AI Institute for Artificial Intelligence and Fundamental Interactions, http://iaifi.org) and by the NSF Graduate Research Fellowship Program under Grant No. 1745302. Any opinions, findings, and conclusions or recommendations expressed in this material are those of the authors and do not necessarily reflect the views of the National Science Foundation. 


\bibliographystyle{splncs04}
\bibliography{lebBib}

\end{document}